\documentclass[
]{ceurart}

\begin{document}

\copyrightyear{2021}
\copyrightclause{Copyright for this paper by its authors.\\
  Use permitted under Creative Commons License Attribution 4.0
  International (CC BY 4.0).}

\conference{In A. Martin, K. Hinkelmann, H.-G. Fill, A. Gerber, D. Lenat, R. Stolle, F. van Harmelen (Eds.), 
Proceedings of the AAAI 2021 Spring Symposium on Combining Machine Learning and Knowledge Engineering (AAAI-MAKE 2021) - 
Stanford University, Palo Alto, California, USA, March 22-24, 2021.}

\title{Deep Learning Approaches for Forecasting Strawberry Yields and Prices Using Satellite Images and Station-Based Soil Parameters}

\author[1]{Mohita Chaudhary}
\ead{m38chaud@uwaterloo.ca}
\address[1]{Department of Electrical and Computer Engineering, University of Waterloo, Ontario, Canada}

\author[1]{Mohamed Sadok Gastli}
\ead{ms2gastli@uwaterloo.ca}

\author[1]{Lobna Nassar}
\ead{lnassar@uwaterloo.ca}

\author[1]{Fakhri Karray}
\ead{karray@uwaterloo.ca}

\setcounter{secnumdepth}{5}

\begin{abstract}
Computational tools for forecasting yields and prices for fresh produce have been based on traditional machine learning approaches or time series modeling. We propose here an alternate approach based on deep learning algorithms for forecasting strawberry yields and prices in Santa Barbara county, California. Building the proposed forecasting model comprises three stages: first, the station-based ensemble model (ATT-CNN-LSTM-SeriesNet\_Ens) with its compound deep learning components, SeriesNet with Gated Recurrent Unit (GRU) and Convolutional Neural Network LSTM with Attention layer (Att- CNN-LSTM), are trained and tested using the station-based soil temperature and moisture data of Santa Barbara as input and the corresponding strawberry yields or prices as output. Secondly, the remote sensing ensemble model (SIM\_CNN-LSTM\_Ens), which is an ensemble model of Convolutional Neural Network LSTM (CNN-LSTM) models, is trained and tested using satellite images of the same county as input mapped to the same yields and prices as output. These two ensembles forecast strawberry yields and prices with minimal forecasting errors and highest model correlation for five weeks ahead forecasts. Finally, the forecasts of these two models are ensembled to have a final forecasted value for yields and prices by introducing a voting ensemble. Based on an aggregated performance measure (AGM), it is found that this voting ensemble not only enhances the forecasting performance by 5\% compared to its best performing component model but also outperforms the Deep Learning (DL) ensemble model found in literature by 33\% for forecasting yields and 21\% for forecasting prices.
\end{abstract}


\maketitle

\section{Introduction}
In Fresh produce Supply Chain Management (FSCM), a crucial part of the procurement process is to find a model which helps in precisely predicting the farmers' prices. These prices are highly affected by the yields hence the availability of accurate yields values to train the forecasting model is crucial \cite{waldick2017integrated}. This study focuses on strawberries as fresh produce, whose yield depends on various parameters related to weather, soil, synthetic factors, irrigation, and others. These factors are quite uncertain therefore the forecasting of strawberry yields and prices is a challenging task. Moreover, from a humanitarian point of view, the United Nations World Food Programme has reported that around 821 million people around the world suffer from hunger \cite{map}, and that number has been growing drastically since the start of the COVID-19 pandemic. This is part of the reason why the United Nations have included ending hunger and the betterment of food security as part of their main goals in their 2030 Agenda for Sustainable Development \cite{unitednations}. A key aspect to overcome these issues and a significant challenge facing food security is the ability to reliably estimate crop yields using forecasting models; the main objective of this work. It should be noted that such forecasting models for strawberries can be applied on numerous similar fresh produces for better yields estimates to sustain food security.

The acquisition of data can be very expensive with limited availability. California is chosen since the data for its strawberry yields, prices, and soil parameters can be acquired from various publicly available online sources. Due to the frequent absence of localized data, using remote sensing data such as satellite images is important since they can cover larger geographic areas. Santa Maria, which lies within Santa Barbara county, is primarily considered in this work because it is one of the largest stations for strawberry produce in California; it is considered the leading state producer for strawberries  \cite{pathak2016evaluating}. The yields and prices are predicted by manipulating historical input data to extract features that can capture as many yields and prices trends as possible. Currently used forecasting tools for fresh produce have limited performance since they do not consider an extensive set of influential factors affecting yields and prices. They are also incapable of capturing the complex patterns in big data sources of prices transactions which provide valuable information on the underlying processes affecting fresh produce prices and quantities; this became feasible by the advent of the state-of-the-art machine learning and DL techniques.

The scope of this work is forecasting both strawberry yields and prices using input features related to soil parameters. The yields and prices values are forecasted 5 weeks ahead. It is found that the past 20 weeks values of the soil parameters affect the yields and hence this is the lag considered to forecast the yields and prices. An aggregated measure is used to gauge the performance of the forecasting models, which are compared to a simple LSTM model and a compound DL ensemble model proposed in \cite{okwuchi2020machine}. The voting regressor ensemble of ATT-CNN-LSTM-SeriesNet\_Ens and SIM\_CNN-LSTM\_Ens models outperforms its compound component models as well as simple DL models such as LSTM. Moreover, it enhances the performance of the compound DL ensemble model described in \cite{okwuchi2020machine} by up to 33\%. 

Section 2 highlights literature review and previous work done in this area, it also summarizes major findings and limitations. The details of the assembled forecasting compound DL models are provided in Section 3. The datasets, data preprocessing, and results of the conducted experiments are presented and analyzed in Section 4. Finally, the drawn conclusion along with future work are found in Section 5.

\section{Literature Review}
Various methods are used for yields prediction like Artificial Neural Networks \cite{karray2004soft,doganis2006time}, K-Nearest Neighbors \cite{maskey2019weather,pathak2016evaluating} and Simple Long Short Term Memory Networks (LSTM) \cite{jiang2018predicting}. The weather parameters are used as input parameters to predict the yields in \cite{kaul2005artificial}; while in \cite{lob,okwuchi2020machine}, the corresponding prices to strawberry yields are predicted using various DL compound models like ConvLSTM, CNN-LSTM, CNN-LSTM-GRU with attention along with DL ensemble models. Adding a self-attention layer improves the yield prediction results to quite an extent. Moreover, the DL models are recommended over nondeep ML and non-ML models for forecasting. In \cite{saad2020machine}, various machine learning and DL imputation techniques for missing values in datasets are discussed. In \cite{jiang2018predicting}, the weather parameters, soil moisture dynamic data and soil quality static data across various counties in Iowa, USA are used as input to forecast the yearly corn yields using a basic LSTM network. The static parameters remain constant over years for a specific region and they are useful only when different regions are considered. The work presented here considers the dynamic parameters solely as input and ignores the static ones since only the Santa Maria county is considered for forecasting yields and prices. In \cite{pantazi2016wheat}, counter-propagation artificial neural networks (CP-ANNs) and Supervised Kohonen Networks (SKNs) are used to predict the wheat yields using a found set of influential soil parameters.
 
As for the use of remote sensing in crop yields forecasting, existing approaches are investigated. Three recent methods and applications are presented in \cite{you2017,joe2018,schwalbert2020}. Their general framework involves using a preprocessed collection of satellite-based data to train neural networks for yields prediction. The first approach is conducted in \cite{you2017} by J. You et al., where they use satellite images that are publicly available to predict annual soybean yields for specific counties in the USA. The images used comprise 7 bands of surface reflectance and 2 bands of temperature. The authors preprocessed these images into histograms due to their large size, then fed them into several prediction models for comparison. The models used are Convolutional Neural Network (CNN) and LSTM. Their results indicate that the CNN models are better than LSTM in predicting yields. However, it should be noted that this approach investigates yearly yields and not daily forecasts. This is further investigated by J. Phongpreecha in \cite{joe2018}, who considers moisture in addition to surface reflectance and temperature satellite images to predict annual corn yields. He also implements dimensionality reduction to histograms as described by \cite{you2017}. The investigated models are a Custom CNN-LSTM, a Separable CNN-LSTM, a CNN-LSTM, a 3D-CNN, and a CNN-LSTM-3D CNN network. The ConvLSTM is found to be the best performing model in forecasting annual corn yields. A similar approach is conducted in \cite{schwalbert2020} where the authors investigate such application in southern Brazil on annual soybean yields. Their approach considers satellite images in addition to precipitation data obtained from weather stations. Three models are tested, namely multivariate OLS linear regression, random forest, and LSTM where, LSTM outperforms the other two models in forecasting. 

From the literature review, it is evident that neural networks, specifically CNN, LSTM, and combinations of both, are best suited for the forecasting application. The limitations of the presented approaches are either considering data collected from localized stations for forecasting, such data is not readily available for all croplands on earth, or giving annual predictions of yields using weekly satellite images. Thus, daily predictions are essential with more complex deep leaning models for higher performance. Furthermore, the satellite images should be used alongside the stationary data to overcome the frequent problem of data scarcity.

\section{Proposed Models and Methodology}
Two models are used for forecasting which are then fed into an ensemble to combine their outputs into a final forecasted value. The first model uses station-based soil data as input while the other relies on soil data captured by satellite images instead. Compound DL models are proved to perform better than simple DL and nondeep ML models as described in \cite{okwuchi2020machine}. The simple DL model, LSTM, and the ensemble model described in \cite{okwuchi2020machine} are compared to the proposed model.

\subsection{The Station-based Model (ATT-CNN-LSTM-SeriesNet\_Ens)}
This model uses the station-based soil data to forecast the strawberry yields and prices. After preprocessing, the data is fed into two compound DL models namely \textbf{Att-CNN-LSTM} and \textbf{SeriesNet with GRU}. The forecasted output values of each of these two models are then fed as input into a voting ensemble which outputs the final forecasted yields or prices.

\subsubsection{Att-CNN-LSTM}
\begin{figure}
  \centering
  \includegraphics[scale=0.7]{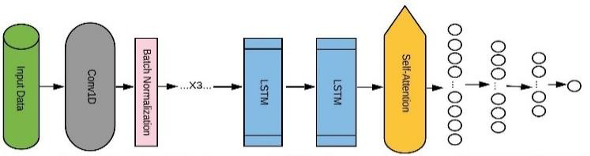}
  \caption{Architecture of CNN-LSTM with Attention}
  \label{fig:Att-ConvLSTM}
\end{figure}

The self-attention layer helps in focusing on the essential details in the input data. In the proposed model, the attention layer uses additive attention and the sigmoid activation function. Figure \ref{fig:Att-ConvLSTM} shows the architecture of the Att-CNN-LSTM compound model. The self-attention is a mechanism which deals with the different positions of a single sequence and then computes the representation of that sequence. The layer is applied atop every unit of the sequence; additive attention is used. The attention function helps in mapping a query and a set of key-value pairs to an output. Here, the keys, output, values and queries are all considered as vectors \cite{vaswani2017attention}.

The Additive Attention works by using a feed forward network to calculate the compatibility function \cite{zhang2019text}. The equations are described in (\ref{ht}), (\ref{et}), (\ref{at}), and (\ref{lt}).
\begin{equation} \label{ht}
    h_{t,t'}=\tanh{(x_t^T W_t +x_{t'}^T W_x + b_t)}
\end{equation}
\begin{equation} \label{et}
    e_{t,t'}=\sigma(W_a h_{t,t'} + b_a)
\end{equation}
\begin{equation} \label{at}
    a_t= \mathrm{softmax}(e_t)
\end{equation}
\begin{equation} \label{lt}
    l_t=\sum_{t'}a_{t,t'}x_{t'}
\end{equation}
where $\sigma$ is the element-wise sigmoid function and $W_x$ and $W_t$ are the weight matrices corresponding to $x_t^T$ and $x_{t'}^T$. $W_a$ is the weight matrix corresponding to the non-linear combination of $W_x$ and $W_t$, while $b_t$ and $b_a$ are the bias vectors \cite{vaswani2017attention}. Equation (\ref{lt}) shows how the attention value $l_t$ is computed. The probability distribution $a_t$ and compatibility score $e_{t,t'}$ must be calculated first to find the value of attention. The compatibility score is calculated using the hidden representation $h_{t,t'}$ of  $x_t^T$ and $x_{t'}^T$. The use of Attention layer contributes significantly to performance by reducing the forecasting error of the DL models. The improvement in performance is evident in the results reported by \cite{ vaswani2017attention, zhang2019text, de2019duo,cai2017cnn} in domains such as natural language processing, fresh produce related predictions and healthcare questionnaires.

Stacking CNNs and LSTMs together helps in utilizing the strength of each of these models \cite{ zheng2019fault, zhang2019text}. The CNNs help in extracting the spatial features whereas the LSTMs help in extracting the temporal features and using their combination highly improves the forecasting performance. The attention layer is added to the CNN-LSTM compound network to improve its performance.

\subsubsection{SeriesNet with GRU}
\begin{figure}
  \centering
  \includegraphics[width=11.5cm,height=7.5cm]{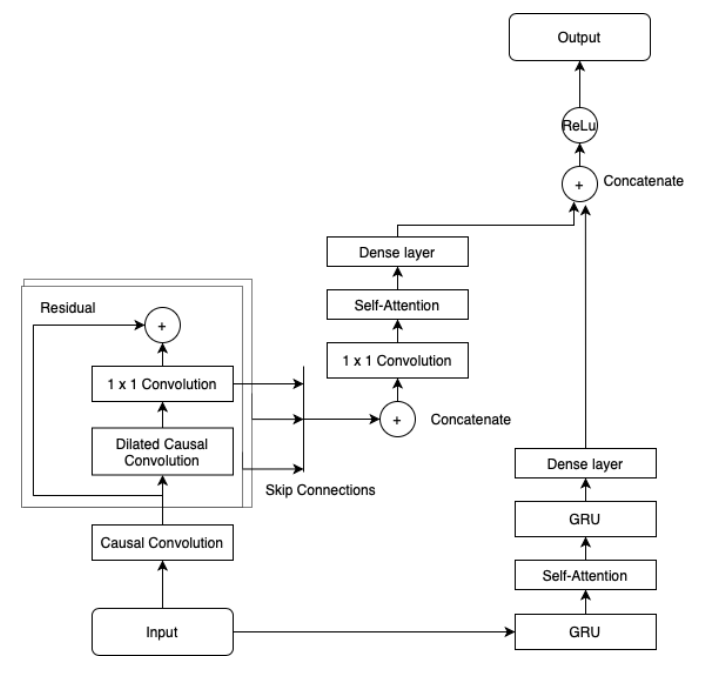}
  \caption{Architecture of SeriesNet with GRU}
  \label{fig:SeriesNetGRU}
\end{figure}

In this model, the GRU is used alongside the dilated causal convolution. The GRU network is a two-layer network with attention module in between. The output at the end is flattened and fed into a single neuron dense layer. The number of layers and neurons alter as per the application. The architecture of the model is illustrated in  Figure \ref{fig:SeriesNetGRU}. Gated Recurrent Unit (GRU) \cite{dey2017gate} is a newer version of RNNs and is quite similar to LSTMs \cite{fu2016using}. GRUs use hidden states rather than using the cell state or memory to transfer the information. They have two gates, a reset gate, and an update gate. The update gate acts similarly to the forget gate and input gates of LSTM. The reset gate, on the contrary, is used to decide how much past information has to be forgotten. GRUs are faster than LSTMs since they have fewer tensor operations. Both LSTMs and GRUs are designed to overcome the short-term memory issues faced by RNNs. Since RNNs face vanishing or exploding gradient issues, GRUs are introduced to mitigate those issues. The structure is similar to those of LSTMs and consists of gates that ensure that the issue of gradient is not encountered.

The traditional time series forecasting models are unable to effectively extract essential data features; thus the authors in \cite{shen2018seriesnet} came up with a novel forecasting architecture called SeriesNet. The SeriesNet consists of two networks, an LSTM network and a dilated causal convolution network. The dilated convolution handles the loss of resolution or coverage due to the down-sampling operation in image semantic segmentation \cite{yu2015multi}. It uses dilated convolutions to systematically aggregate multi-scale contextual information and improve the accuracy of image recognition. The causal convolution ensures that the convolution kernel of CNN can perform convolution operations exactly in time sequence \cite{oord2016wavenet}, and that the convolution kernel only reads the current and historical information. The LSTM network aims to learn holistic features and to reduce dimensionality of multi-conditional data. The combined results of the networks help the models to learn multi-range and multilevel features from time series data, hence it has higher predictive accuracy compared to other models. The SeriesNet with GRU model uses residual learning as well as batch normalization to improve generalization.

\subsection{The Remote Sensing Ensemble Model (SIM\_CNN-LSTM\_Ens)}
\begin{figure}
\centering
\begin{minipage}{.5\textwidth}
  \centering
  \includegraphics[width=1.0\linewidth]{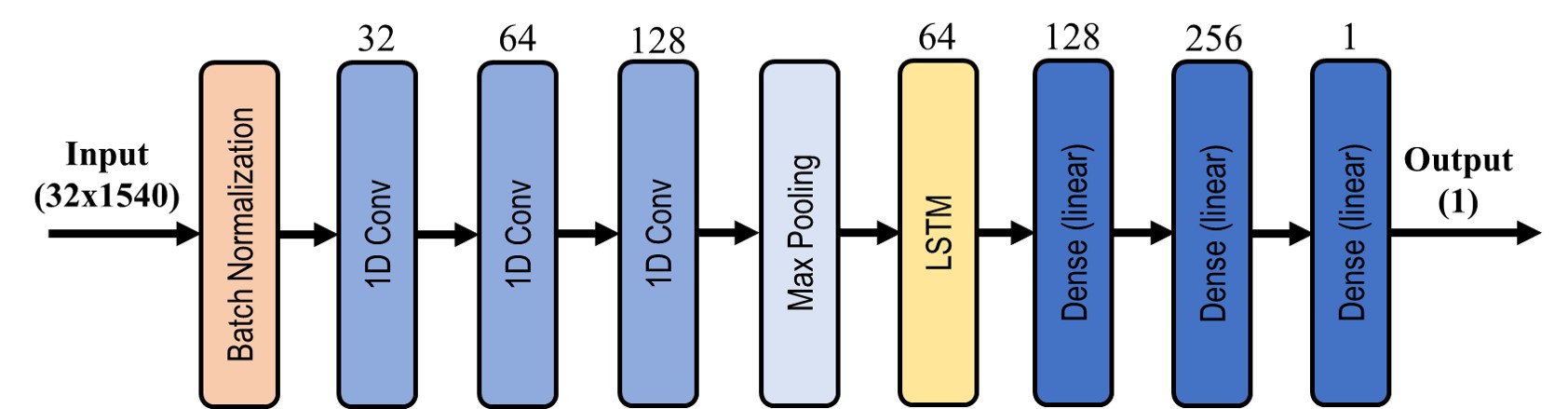}
  \captionof{figure}{CNN-LSTM architecture for yields \newline forecasting using satellite images}
  \label{fig:CNNLSTM_yield}
\end{minipage}%
\begin{minipage}{.5\textwidth}
  \centering
  \includegraphics[width=1.0\linewidth]{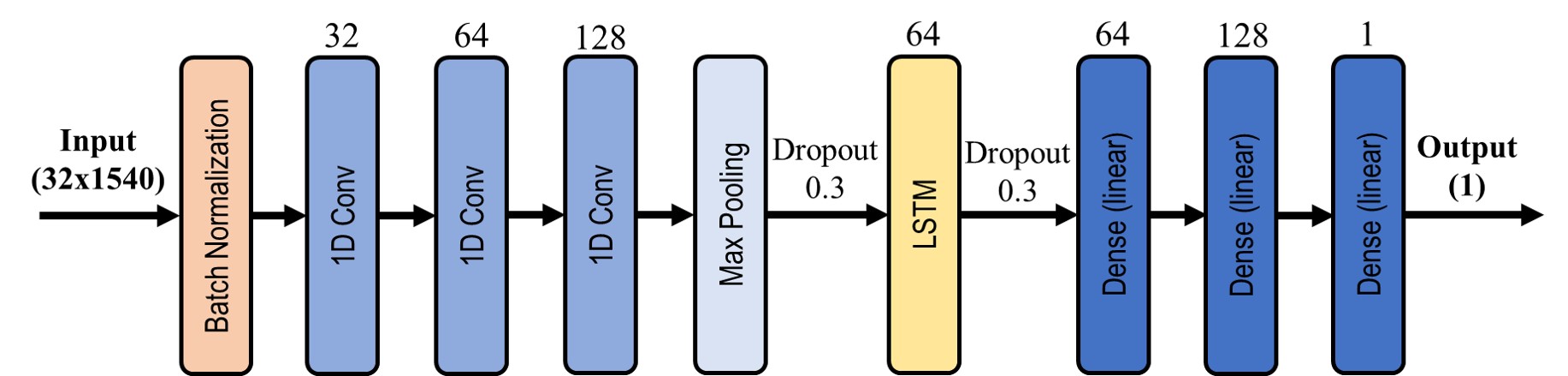}
  \captionof{figure}{CNN-LSTM architecture for prices \newline forecasting using satellite images}
  \label{fig:CNNLSTM_price}
\end{minipage}
\end{figure}

This model is an ensemble of CNN-LSTM. It incorporates the capabilities of both CNN and LSTM by treating them as layers. The architecture for the CNN-LSTM model used for yields forecasting is presented in Figure \ref{fig:CNNLSTM_yield}, which differs from the prices forecasting model. It should also be noted that the models are fine tuned through trial and error of multiple configurations. As for the CNN-LSTM prices forecasting model, its architecture is presented in Figure \ref{fig:CNNLSTM_price}. The preprocessing is implemented using the Geospatial Data Abstraction Library (GDAL) \cite{gdal} in Python to import the images into a processable format. After preprocessing, the features are fed into the designed models. These models are implemented uisng Python's Tensoflow Keras library \cite{chollet2015keras} due to its user-friendliness.

\subsection{Ensemble of ATT-CNN-LSTM-SeriesNet\_Ens and SIM\_CNN-LSTM\_Ens}
\begin{figure}
  \centering
  \includegraphics[scale=0.4]{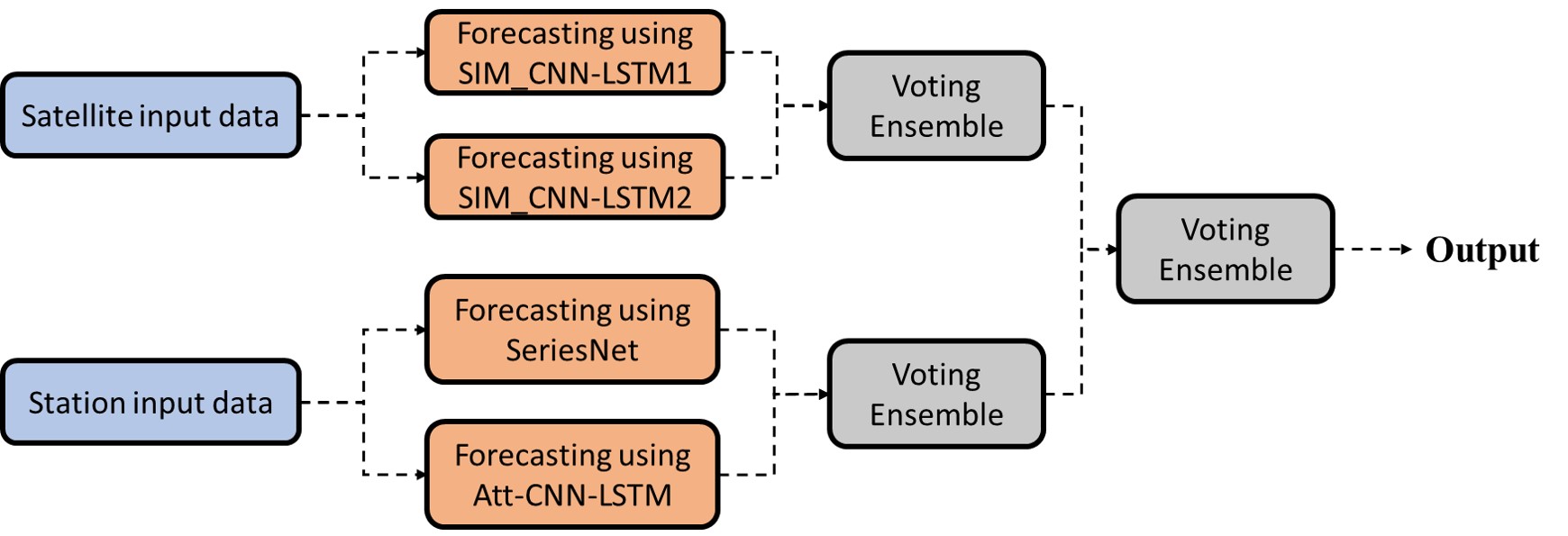}
  \caption{Block diagram of the proposed ensemble model}
  \label{fig:blockdiagram}
\end{figure}

The previous approaches extract different features from their data, therefore averaging their forecasted outputs could potentially lead to a better overall forecast. Hence, a voting ensemble is required to achieve this averaging. Figure \ref{fig:blockdiagram} illustrates the overall architecture of the final proposed model. The outputs of both compound models are combined by the averaging ensemble to obtain the final forecast.

\subsection{Evaluation Metrics}
The metrics used for analysis are the Mean Absolute Error (MAE), the Root Mean-Squared Error (RMSE), R-Squared coefficient ($R^2$), and the Aggregated Measure (AGM). The unit in which MAE is measured is pounds/acre for yield and US dollars for price. The AGM is a measure composed of all the previously mentioned metrics, whose purpose is to attempt to incorporate the information captured by all three metrics into one metric to simplify the process of deciding the best performing model \cite{lob,nassar2020imputation}. The measure is negatively-oriented, meaning lower scores indicate better performance. It is mathematically defined in (\ref{agm}).
\begin{equation} \label{agm}
AGM = \frac{RMSE + MAE}{2} \times (1 - R^2)
\end{equation}

\section{Datasets, Experiments and Analysis}
\subsection{Datasets and Preprocessing}
\subsubsection{Station-based Data}
There are various parameters related to the soil affecting the yields and prices of fresh produce namely: soil moisture, soil temperature, solar radiation, surface temperature, PDSI (Palmer Drought Severity Index) \cite{w11,w10},... etc. Choosing the most influential parameters is a major challenge due to the high correlation amongst those parameters. Hence, the Random Forest feature selection method in Python with scikit-learn is used and the soil moisture and temperature are selected accordingly as the most important parameters. Moreover, soil moisture and temperature are the two soil parameters that can be obtained from satellite images as well. The station-based soil data is downloaded from the National Oceanic and Atmospheric Administration website \cite{wnoc}. The data for strawberry yields and prices are downloaded from the California Strawberry Commission website \cite{w1}. A lag of the past 20 weeks, i.e. 140 days, of soil parameters values is found to affect the yields forecasting and prices values 5 weeks ahead. Two daily parameters are considered, soil temperature and moisture, hence the total number of input parameters adds up to 280 (2 parameters x 140 days). After normalizing the data, the Principal Component Analysis (PCA) \cite{abdi2010principal} is applied on the 280 parameters and it is found that the first 36 parameters gave the maximum proportion of variance therefore are chosen to train and test the forecasting model along with the corresponding yields or prices output. The total number of available samples is 2812 (from year 2011 to 2019) out of which 80\% are used for training and 20\% for testing.

\subsubsection{Remote Sensing Data}
\begin{figure}
\centering
\begin{minipage}{0.5\textwidth}
  \centering
  \includegraphics[ width=1.0 \linewidth,,height=4.05cm]{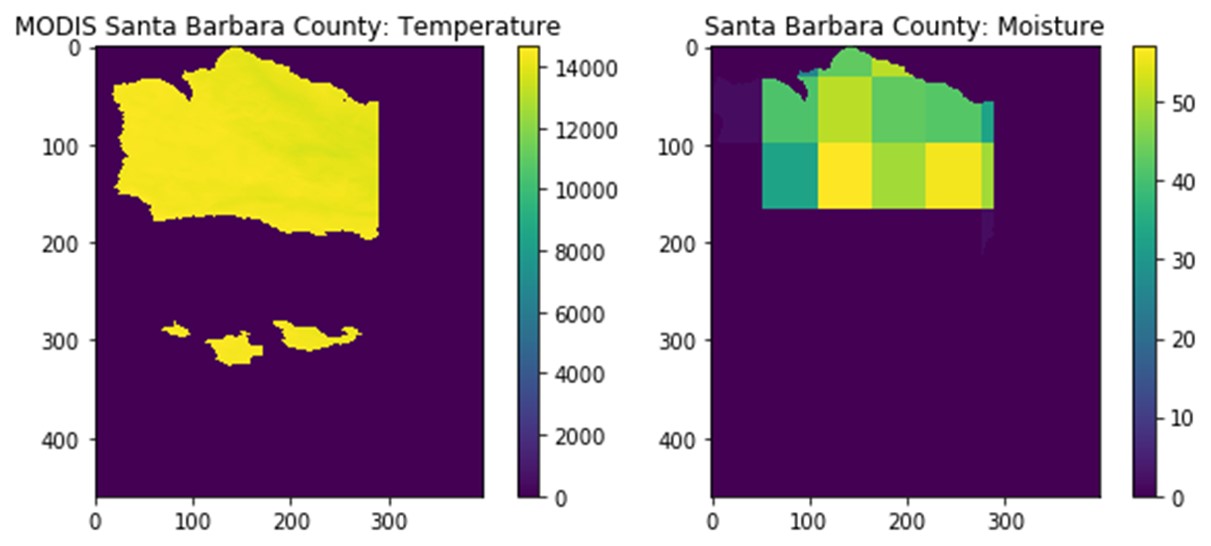}
  \captionof{figure}{ \newline Image samples from the temperature\newline bands (left) and the moisture bands (right)}
  \label{fig:satImages}
\end{minipage}%
\begin{minipage}{0.5\textwidth}
  \centering
  \includegraphics[width=1.1 \linewidth,height=5cm]{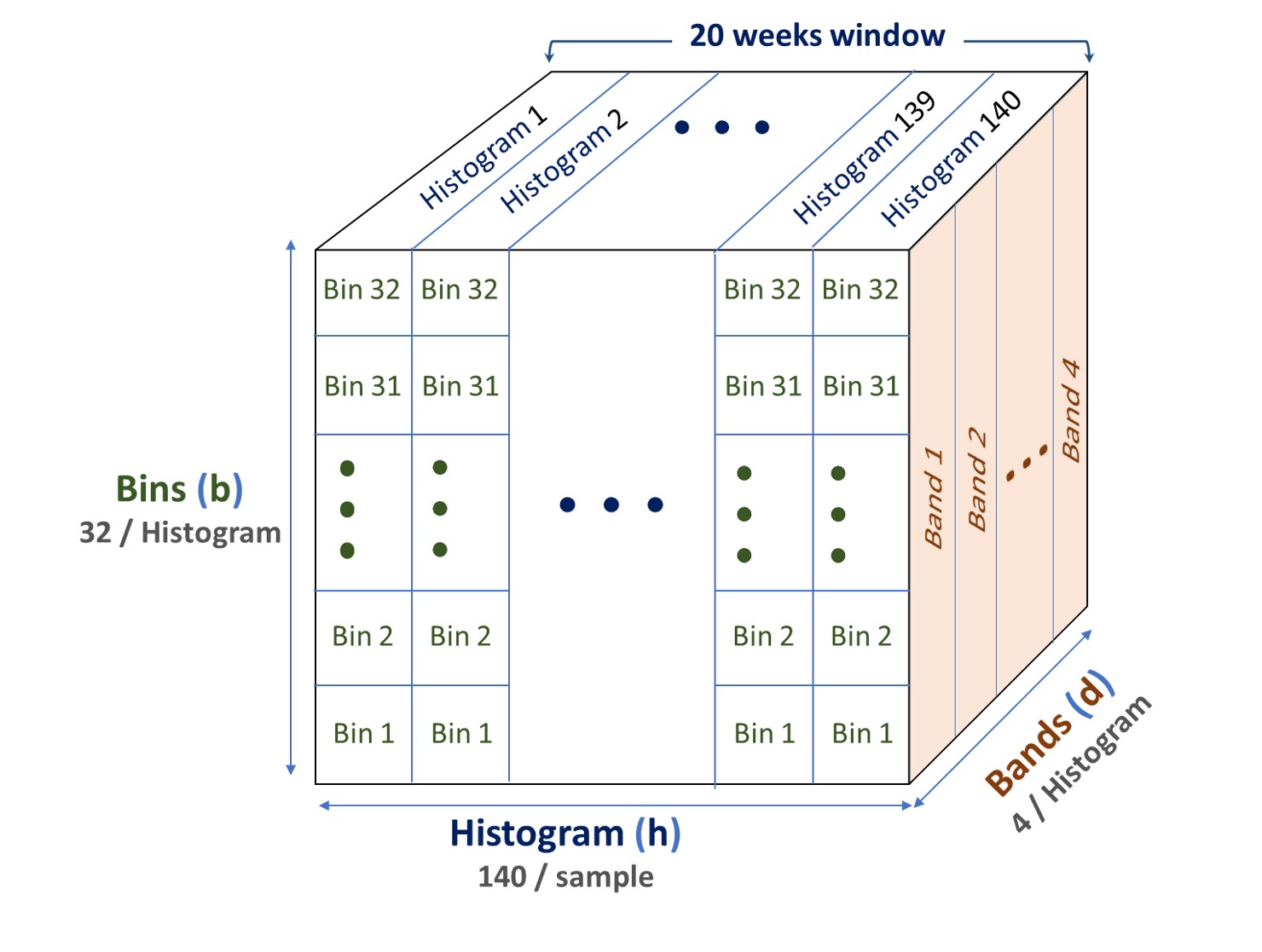}
  \captionof{figure}{Input sample histogram structure}
  \label{fig:Cube}
\end{minipage}
\end{figure}

The remote sensing data contains two sets of satellite images; one for surface temperature data and the other for moisture data. The images in the temperature dataset, obtained from \cite{sr_temp}, are taken daily, whereas those in the moisture dataset, obtained from \cite{moist}, are taken every 3 days. Therefore, each moisture image is duplicated twice, once for the day before the original image and once for the day after, in order to have daily moisture values along with the daily temperature. This approach is an approximation of the missing days based on the closest known data to those days. Samples of the original images obtained from the datasets for Santa Barbara county are presented in Figure \ref{fig:satImages}. The moisture bands are for the mainland moisture levels, which is why the data is not available for the islands and coastal areas. In addition, a land cover mask is applied to all images, obtained from the MODIS database \cite{sr_temp}. The mask maintains the pixel values of the image parts of interest, while setting pixels not within the mask to a zero value, to ensure that the model only trains on pixel values that correspond to crop lands hence minimizing the number of pixels processed. The total number of samples is 2977 (from year 2011 to 2019) out of which 80\% are used for training and 20\% for testing.

As shown in \cite{you2017}, the images are too large to be fed directly into the model. Thus, dimensionality reduction is applied by converting them into histograms of pixel frequency counts. This reduction is based on the assumption of permutation invariance, meaning that shuffling the pixels has no effect on the information retained in the image. The histograms are then normalized based on their pixel values. Surface reflectance bands used in \cite{you2017} and \cite{joe2018} are omitted due to their negligible improvement to the prediction performance for the considered county and fresh produce. The number of bins, $b$, is set to be 32; as it is a reasonable spread of pixel counts as found in \cite{you2017}. Moreover, the range of pixel values for each band is chosen to maximize the spread of the pixels count distribution. The structure is visualized in Figure \ref{fig:Cube}; where the three axes represent the three dimensions of the histogram. Before being fed to the models, the bins and bands dimensions are flattened into a single dimension to fit the models.

\subsection{Experiments and Analysis}

The block diagram in Figure \ref{fig:blockdiagram} depicts the proposed final model, which has a voting ensemble which averages the forecasted yields or prices resulting from the two proposed DL compound ensembles described in Section 3.1 and Section 3.2. The main idea behind this averaging is that the final forecast should pick the trends forecasted by both of the component models.

\begin{figure}
\centering
\begin{minipage}{0.45\textwidth}
  \centering
  \includegraphics[width=0.8\linewidth]{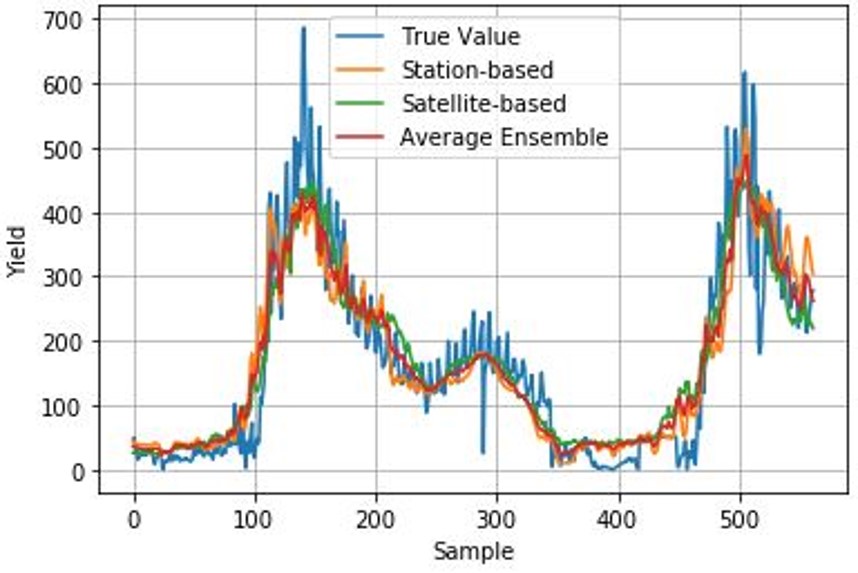}
  \captionof{figure}{Forecasted vs. true yields values}
  \label{fig:MT2Y}
\end{minipage}%
\begin{minipage}{0.45\textwidth}
  \centering
  \includegraphics[width=0.8\linewidth]{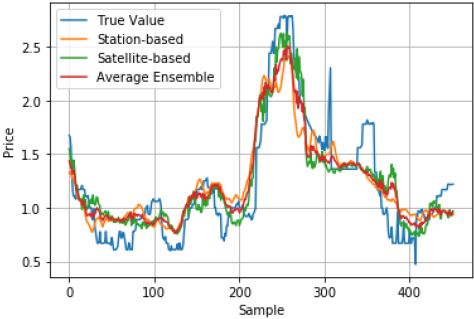}
  \captionof{figure}{Forecasted vs. true prices values}
  \label{fig:MT2P}
\end{minipage}
\end{figure}

For yield forecasting, Figure \ref{fig:MT2Y} shows the true yields versus the yields forecasted by ATT-CNN-LSTM-SeriesNet\_Ens, SIM\_CNN-LSTM\_Ens, and their averaging ensemble. The figure shows that the voting ensemble successfully follows the general trend of the true values despite its inability to describe the sharp fluctuations. Table \ref{table:MT2Y_ens} presents different metrics scores of the forecasted values for the different models. Although SIM\_CNN-LSTM\_Ens outperforms ATT-CNN-LSTM-SeriesNet\_Ens in AGM, the averaging ensemble outperforms both of its weaker component models with its lowest AGM. This indicates that there is some information obtained by the latter that enhances the performance compared to the individual models. Based on AGM, the voting ensemble enhances the forecasting performance over that of the DL ensemble model in \cite{okwuchi2020machine} by 33\% and that of the simple LSTM DL by 56\%. 

For prices forecasting, Figure \ref{fig:MT2P} depicts the true prices versus the prices forecasted by ATT-CNN-LSTM-SeriesNet\_Ens, SIM\_CNN-LSTM\_Ens, and the averaging ensemble of both. Table \ref{table:MT2P} presents different performance metrics scores for the tested models. Although ATT-CNN-LSTM-SeriesNet\_Ens outperforms SIM\_CNN-LSTM\_Ens with a noticeable AGM difference, the voting ensemble outperforms both of these compound components with its lowest AGM. The voting ensemble persists to be the highest performing compared to the LSTM model, 53\% improvement, and to the ensemble model in \cite{okwuchi2020machine}, 21\% improvement in AGM.

\begin{table*}
  \caption{Results for using LSTM, SIM\_CNN-LSTM\_Ens, and ATT-CNN-LSTM-SeriesNet\_Ens to forecast yield}
  \label{table:MT2Y_ens}
  \smallskip\noindent
\resizebox{\linewidth}{!}{
  \begin{tabular}{cccccc}
    \toprule
    Score & LSTM & Ensemble in \cite{okwuchi2020machine} & SIM\_CNN-LSTM\_Ens  & ATT-CNN-LSTM-SeriesNet\_Ens & Voting Ensemble\\
    \midrule
    MAE & 53.1 & 42.5 & 39.1 & 40.7 & 37.0\\
    RMSE & 70.8 & 62.2 & 55.2 & 58.8 & 54.6\\
    $R^2$ & 0.780 & 0.83 & 0.866 & 0.848 & 0.869\\
    AGM & 13.6 & 9.0 & 6.3 & 7.5 & 6.0\\
    \bottomrule
  \end{tabular}}
\end{table*}

\begin{table*}
  \caption{Results for using LSTM, SIM\_CNN-LSTM\_Ens, and ATT-CNN-LSTM-SeriesNet\_Ens to forecast price}
  \label{table:MT2P}
    \smallskip\noindent
\resizebox{\linewidth}{!}{
  \begin{tabular}{cccccc}
    \toprule
    Score & LSTM & Ensemble in \cite{okwuchi2020machine} & SIM\_CNN-LSTM\_Ens & ATT-CNN-LSTM-SeriesNet\_Ens & Voting Ensemble\\
    \midrule
    MAE & 0.268 & 0.21 & 0.227 & 0.214 & 0.208\\
    RMSE & 0.341 & 0.27 & 0.292 & 0.263 & 0.264\\
    $R^2$ & 0.609 & 0.72 & 0.712 & 0.766 & 0.764\\
    AGM & 0.119 & 0.07 & 0.0748 & 0.0557 & 0.0555\\
    \bottomrule
  \end{tabular}}
\end{table*}

\section{Conclusion and Future Work}
This paper explores the application of multiple DL models in forecasting strawberry yields and prices. Station-based data is used to train the ATT-CNN-LSTM-SeriesNet\_Ens, while remote sensing-based data is used to train the SIM\_CNN-LSTM\_Ens for forecasting. It is found that the SIM\_CNN-LSTM\_Ens model does better at yields forecasting while the ATT-CNN-LSTM-SeriesNet\_Ens model is better at prices forecasting. The voting ensemble of these models outperforms its individual components since each component provides a different forecasting behavior. Moreover, it further proves to perform better by up to 33\% compared to the most recent DL ensemble forecasting model. An acknowledged limitation is the restricted ability of the deployed models to capture steep fluctuations in yields and prices, making them not predictable with the available tools. Moreover, having an ensemble of various compound DL models is computationally expensive. Potential future work is needed to further improve the ability of the models to capture steep fluctuations in both yields and prices. Generalization of the models application to other fresh produces is also required; this is achieved through transfer learning to forecast yields and prices of FPs similar to strawberries efficiently with minimal retraining.

\begin{acknowledgments}
The authors would like to acknowledge the financial support provided by Loblaws corporation, NSERC CRD program and Mitacs.
\end{acknowledgments}

\bibliography{aaai-make}


\end{document}